# A Dynamic Time Warping-Transfer Learning Approach to Transferring Knowledge in Stress-strain Behaviors from Polymers to Metals: An Affordable and Generalizable Additive Manufacturing Part Qualification Framework

Chenglong Duan[1] and Dazhong Wu[1,*]

[1] Department of Mechanical and Aerospace Engineering, College of Engineering and Computer Science, University of Central Florida, Orlando, FL 32816, USA

* Corresponding author
Email: dazhong.wu@ucf.edu

**Abstract**

Part qualification in additive manufacturing (AM) ensures that additively manufactured parts can be consistently produced and reliably used in critical applications. One crucial aspect of part qualification is to determine the complex stress-strain behavior of additively manufactured parts. However, conventional part qualification techniques such as the destructive testing and non-destructive testing are costly and time consuming, especially for metal AM. To address this challenge, we develop a dynamic time warping (DTW)-transfer learning (TL) framework for AM part qualification by transferring knowledge gained from the stress–strain behaviors of additively manufactured low-cost polymers to high-performance, expensive metals. Specifically, the framework selects one single optimal polymer dataset that is the most similar to the metal dataset in the target domain using DTW among multiple polymer datasets, including Nylon, PLA, CF-ABS, and Resin. A long short-term memory (LSTM) model is then trained on one single optimal polymer dataset and tested on one of three target metal datasets, including AlSi10Mg, Ti6Al4V, and carbon steel datasets. Experimental results show that the Resin dataset is selected as the optimal polymer dataset in the source domain for the AlSi10Mg and Ti6Al4V datasets, while the Nylon dataset is selected as the optimal polymer dataset in the source domain for the carbon steel dataset. The DTWTL model trained on one single optimal polymer dataset as the source domain achieves the best predictive performance, including an average mean absolute percentage error of 12.41%, an average root mean squared error of 63.75, and an average coefficient of determination of 0.96 when three metals are used as the target domain, outperforming the vanilla LSTM model without TL as well as the TL model trained on all four polymer datasets as the source domain.

## 1. Introduction

Additive manufacturing (AM) has a wide range of applications in aerospace, automative, energy, defense, and medical industries [1-7]. Metal AM, in particular, is crucial to critical applications due to the superior mechanical properties of additively manufactured metal parts [8-10]. However, metal AM has not been widely adopted in critical industries due to the defects induced by metal AM, uncertainty in the mechanical performance, and the high cost and lengthy process of testing and inspecting the additively manufactured metal parts [11-13]. Therefore, an affordable and generalizable part qualification framework is crucial for metal AM because it verifies that additively manufactured metal parts satisfy the required design specifications to ensure these parts can be consistently reproduced, scaled for mass production, and reliably used in critical applications [14, 15]. Part qualification usually involves predicting the defects and mechanical performance of additively manufactured parts. Determining the complex stress-strain behavior of additively manufactured metal parts is one of the critical aspect in part qualification [16, 17]. However, conventional part qualification techniques used for determining the stress-strain behavior involve costly, time-consuming, and labor-intensive destructive and non-destructive testing (NDT). While destructive testing provides highly accurate data, expensive additively manufactured metal parts must be destroyed in physical testing. NDT techniques inspect additively manufactured metal parts without causing damage by in-process and post-process measurement techniques such as sensor-based in-process monitoring as well as ultrasound and computed tomography (CT) scanning. However, NDT techniques are costly and time consuming. For example, in-process monitoring often requires expensive measurement techniques such as thermal imaging or in-situ X-ray.

With the recent advances in artificial intelligence (AI) and machine learning (ML), data-driven methods have been increasingly used to predict the stress-strain behavior of additively manufactured parts with high accuracy and efficiency [18-24]. While conventional neural networks are widely used to predict the stress-strain behavior, other deep learning techniques such as long short-term memory (LSTM) and temporal convolutional network (TCN) have been increasingly used to predict the stress-strain behavior. Meanwhile, these ML models have leveraged diverse input features, including part geometry, AM process parameters, microstructural descriptors, and in-situ monitoring data [25-27]. However, one research gap is that training an effective ML model for additively manufactured metal parts often requires large volumes of labeled data, which is costly and time consuming to create [28, 29]. Specifically, the raw metal powders required by laser-based metal AM can cost around $90 to $600 per kilogram. Moreover, fabricating parts via metal AM and conducting destructive testing involve significant operational costs. Therefore, the development of an affordable part qualification framework for metal AM is critical to the AM industry.

Another research gap is that most of the conventional ML models are not generalizable when materials, part geometry, and process conditions vary. Transfer learning (TL) models have the potential to build a generalizable model with limited data [30, 31] to predict the mechanical performance of additively manufactured parts [32-35] and defects [36-39]. To the best of our knowledge, very few studies have been reported on predicting the stress-strain behavior of additively manufactured metal parts using TL. In a prior study conducted by Yang et al. [40]. In this study, a cross-material sequence-to-sequence framework was introduced to improve predictive performance for aluminum alloys with limited data by leveraging knowledge gained from high-strength steel data. However, transferring knowledge within metallic systems is still not an affordable solution for metal AM because of the high costs of metal powder and machinery. One potential solution is to predict the stress-strain behavior of additively manufactured metals by leveraging knowledge gained from additively manufactured low-cost materials such as polymers because common polymer filaments cost only $20 to $50 per kilogram.

Moreover, in the context of TL, to pre-train a TL model cost-efficiently with limited data in the source domain, one challenge is to determine whether one single optimal dataset or multiple datasets in the source domain should be used to pre-train a TL model. In the context of this study, the challenge is to determine whether one single polymer dataset or multiple polymer datasets in the source domain should be used to pre-train a TL model. The most common strategy is to simply use all the datasets available in the source domain to build a multi-source TL model [41, 42]. However, increasing the data size in the source domain does not necessarily improve the predictive performance [34, 43]. In fact, adding more training data in the source domain that is less similar to the target domain can result in poor predictive performance [44-46]. Therefore, selecting one single optimal source domain dataset that is the most similar to the target domain from multiple source domain datasets is crucial to the predictive performance of a TL model. To address this issue, dynamic time warping (DTW), a technique that measures similarity between time-series data [47, 48], can be used to measure the similarity of stress-strain curves between polymers and metals. By capturing similarities in stress-strain curves that may vary significantly with material type, DTW can select one single optimal polymer dataset in the source domain that is the most similar to a metal dataset in the target domain.

To address the aforementioned issues, a novel DTW-TL framework was developed to predict the stress-strain behaviors of additively manufactured high-performance, expensive metals by leveraging knowledge gained from additively manufactured low-cost polymers. The DTW-TL framework consists of four major steps. First, the DTW algorithm is used to identify one single optimal polymer dataset in the source domain that exhibits the most similar stress-strain behavior (i.e., the minimum DTW distance) to a metal dataset in the target domain. Second, an LSTM model is pre-trained on the single optimal polymer dataset. Third, the pre-trained LSTM model is fine-tuned on the training dataset in the target domain. Finally, the fine-tuned LSTM model is tested on the test dataset in the target domain. Specifically, the source domain contains data collected from four polymers, including Nylon (Ultimaker), polylactic acid (PLA, MatterHackers), carbon fiber-acrylonitrile butadiene styrene (CF-ABS, Push Plastic), and photopolymer resin (Resin, Flashforge). Nylon, PLA, and CF-ABS samples were fabricated by fused filament fabrication (FFF) and Resin samples were fabricated by digital light processing (DLP). The target domain contains data collected from three metals, including AlSi10Mg (3D Systems, Inc), Ti6Al4V (3D Systems, Inc), and carbon steel (Böhler welding). AlSi10Mg and Ti6Al4V samples were fabricated by laser powder bed fusion (L-PBF), and carbon steel samples were fabricated by wire arc additive manufacturing (WAAM). Experimental results showed that the DTW-TL model trained on one single optimal polymer dataset in the source domain selected among four polymer datasets achieved the best predictive performance compared to the vanilla LSTM model without TL and the TL model trained on all four polymer datasets as the source domain. The main contributions of this study are summarized as follows:

(1) A generic dynamic time warping-transfer learning framework was developed to predict the stress-strain behaviors of additively manufactured high-performance, expensive metals by transferring knowledge from additively manufactured low-cost polymers.
(2) A dynamic time warping-based source domain selection method was developed to quantitatively evaluate the similarity between the stress-strain behaviors of source and target domain datasets as well as select one single optimal polymer dataset in the source domain dataset which is the most similar to the metal dataset in the target domain.
(3) The effectiveness of the dynamic time warping-transfer learning framework was demonstrated on a comprehensive dataset including four additively manufactured polymers and three additively manufactured metals.

## 2. Material and methods

### 2.1 Transfer learning framework

Fig. 1 illustrates the DTW-TL framework for predicting the stress-strain behaviors of additively manufactured high-performance, expensive metals by leveraging knowledge gained from additively manufactured low-cost polymers. First, as shown in Fig. 1a, the stress-strain curves ($\varepsilon$ denotes for strain and $\sigma$ denotes for stress) of multiple polymers (i.e., Nylon, PLA, CF-ABS, and Resin) in the source domain are compared with the stress-strain curves of multiple metals (i.e., AlSi10Mg, Ti6Al4V, and carbon steel) in the target domain training dataset via DTW to identify one single optimal source polymer whose stress-strain behavior is the most similar to the target metal (i.e., minimum DTW distance). The single optimal polymer dataset is then used to pre-train a LSTM model. Specifically, as shown in Fig. 1b, the polymer dataset is split into an input matrix and a corresponding output vector. Each input matrix consists of a sequence of strain values ($\varepsilon^t$, time-variant) along with constant process parameters ($P$, time-invariant) concatenated at each time step. Each output vector consists of the stress values at each time step ($\sigma^t$, time-variant).

Second, the parameters of the pre-trained model are transferred to another LSTM model, which is then re-trained on the stress-strain data and corresponding process parameters of the target domain training dataset. In this study, the parameter transfer strategy is used because it allows one to retrain the pre-trained model by leveraging knowledge gained from the source domain. Fine-tuning adjusts the model parameters to better fit the target domain data, often resulting in faster convergence and improved predictive performance, particularly when the size of the target dataset is limited. In the parameter transfer process, the model parameters of the pre-trained LSTM are denoted as $\theta_s$. For the target model, we initialize its parameters $\theta_t$ with those from the source model as follows:

$$\theta_t^0 = \theta_s \tag{1}$$

The target model is then fine-tuned using the training data of the target domain by minimizing a loss function $L_t$, this fine-tuning process is iteratively performed as follows

$$\theta_t^{k+1} = \theta_t^k - \eta \nabla_\theta L_t(\theta_t^k) \tag{2}$$

where $\theta_t^k$ represents the target model parameters at iteration $k$, $\eta$ is the learning rate, and $\nabla_\theta L_t(\theta_t^k)$ is the gradient of the loss function with respect to the parameters. Finally, in Fig. 1d, the fine-tuned LSTM model is validated on the test dataset in the target domain. It should be noted that Fig. 1 shows the DTW-TL framework that predicts the stress-strain behaviors of the AlSi10Mg samples as an example. The framework for predicting the stress-strain behaviors of the Ti6Al4V and carbon steel samples is the same.

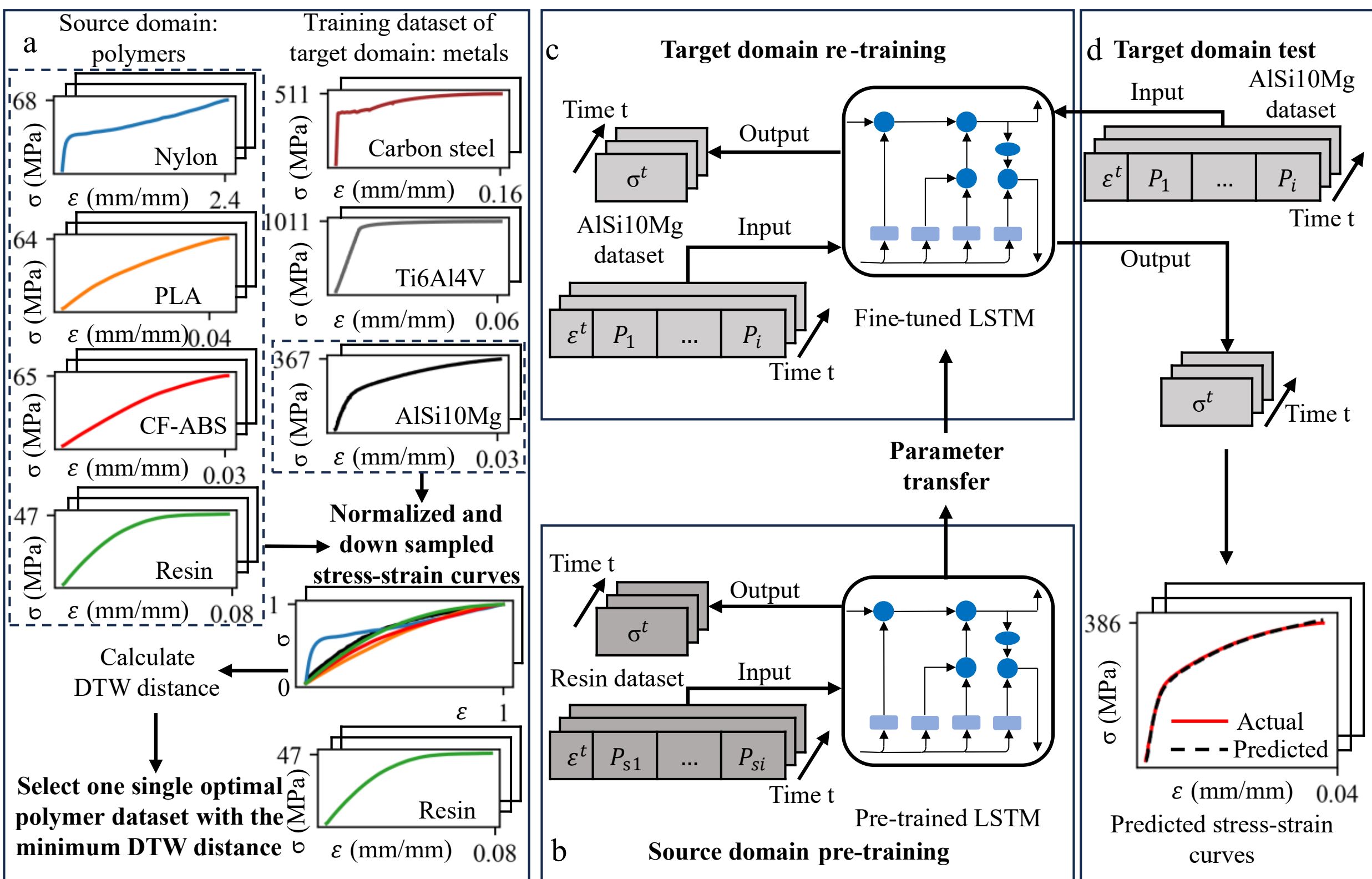

Fig. 1. Overview of the DTW-TL framework. (a) Selecting one single optimal source domain polymer dataset using DTW. (b) Pre-training an LSTM model on one single optimal source domain dataset. (c) Re-training the pre-trained LSTM model on the target domain training dataset. (d) Predicting the stress-strain behaviors of the target domain test dataset.

For the LSTM model built in this work, it was originally developed from the recurrent neural network (RNN) [49]. By incorporating a gate-controlled memory cell, LSTM is designed to learn the long-term temporal dependencies in sequential data. Fig. 2a shows a layered LSTM architecture with an input layer, an LSTM layer consisting of multiple LSTM cells, a flatten layer, a fully connected layer, and an output layer. At each time step $x^t$, the input flows into an LSTM cell, enabling the network to learn temporal dependencies. The outputs of the LSTM layer then pass into a flatten layer, which converts the data into a one-dimensional vector to feed into the fully connected layer. Finally, this fully connected layer yields predictions at each time step, forming the output layer.

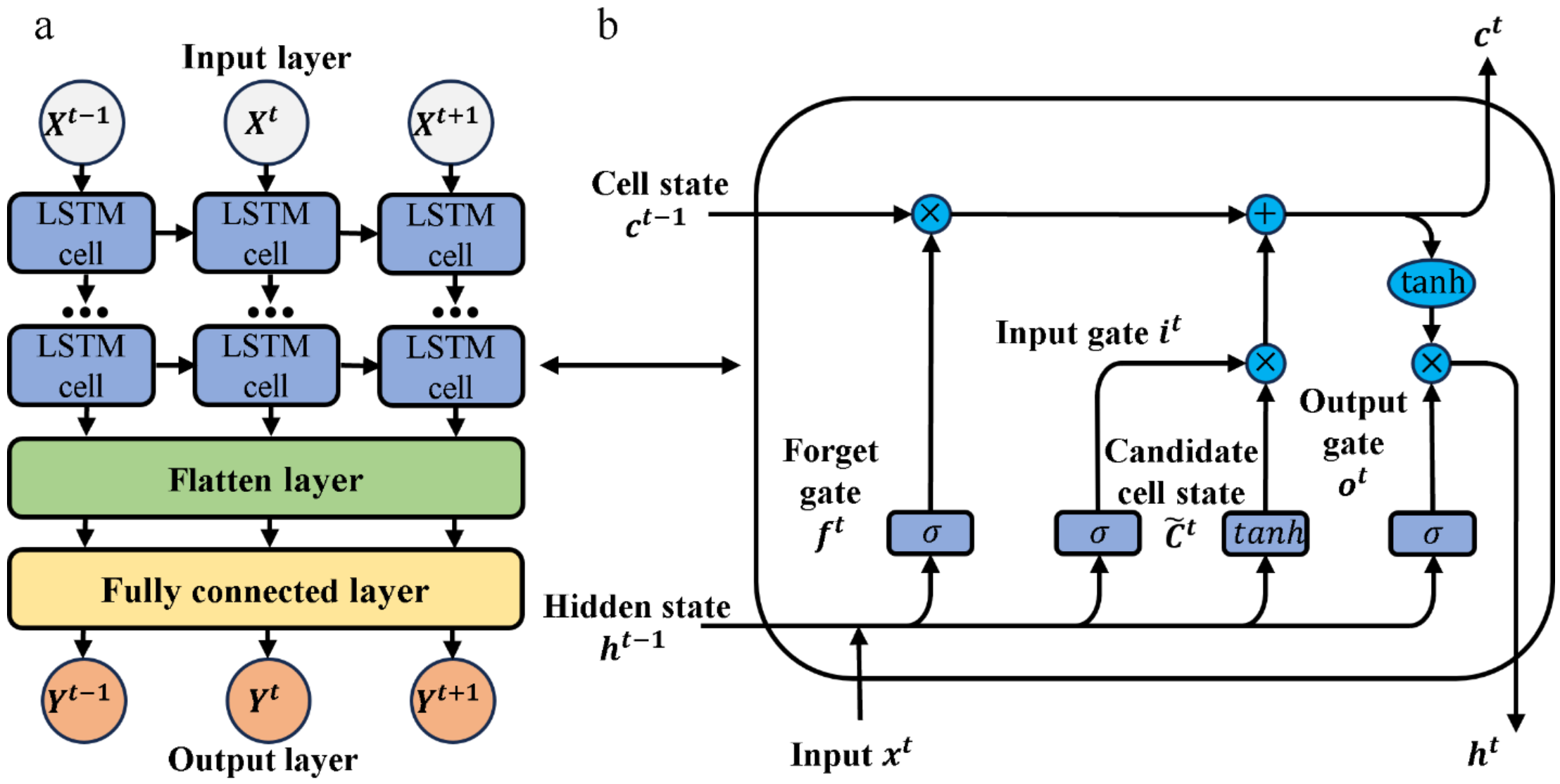

Fig. 2. Architecture of a typical LSTM model. (a) A layered LSTM model. (b) Structure of the LSTM cell.

Fig. 2b illustrates the internal architecture of a single LSTM cell, highlighting how it mitigates the vanishing or exploding gradient problems in RNNs. The core component is the cell state $c^{t}$, which acts as a memory capable of storing information over extended time steps. Three gates include forget gate $f^{t}$, input gate $i^{t}$, and output gate $o^{t}$, controlling the flow of information into and out of the cell state. The forget gate determines how much of the previous cell state $c^{t-1}$ should be retained, while the input gate decides the degree to which new candidate information $\tilde{C}^{t}$ updates the current cell state. Finally, the output gate governs how much of the updated cell state is revealed as the hidden state $h^{t}$ at the current time step. Moreover, in the LSTM cell diagram, symbols $\otimes$ and $\oplus$ indicate pointwise multiplication and addition, respectively, while the $\sigma_{sig}$ represent the sigmoid activation and the $tanh$ block represents the hyperbolic tangent function. Mathematically, the forward-propagation procedure for an LSTM with a forget gate can be expressed as in Eqs. (3)-(8), where $W_{fh}$, $W_{fx}$, $W_{ih}$, $W_{ix}$, $W_{ch}$, $W_{cx}$, $W_{oh}$, and, $W_{ox}$ are the weight matrices for each of the gates, and $b_{f}$, $b_{i}$, $b_{c}$, $b_{o}$ are their respective biases. These learnable parameters allow the network to adapt its memory mechanism to the specific temporal patterns in training data, effectively capturing long-term dependencies during backpropagation.

Forget gate $f^{t} = \sigma_{sig}(W_{fh}h^{t-1} + W_{fx}x^{t} + b_{f})$ (3)

Input gate $i^{t} = \sigma_{sig}(W_{ih}h^{t-1} + W_{ix}x^{t} + b_{i})$ (4)

Candidate cell state $c^{t} = f^{t} * c^{t-1} + i^{t} * \tilde{C}^{t}$ (5)

Memory cell state $\tilde{C}^{t} = \tanh\,(W_{ch}h^{t-1} * W_{cx}x^{t} + b_{c})$ (6)

Output gate $o^{t} = \sigma_{sig}(W_{oh}o^{t-1} + W_{ox}x^{t} + b_{o})$ (7)

Hidden state $h^{t} = o^{t} * \tanh\,(c^{t})$ (8)

In this study, the LSTM model was built using PyTorch and was trained on a computer equipped with an NVIDIA GeForce RTX 4090 GPU, an Intel i9 14900KF CPU, and 32GB of RAM. The model consists of one LSTM layer that contains 32 hidden units, followed by a fully connected layer that maps the final hidden state to a single stress as the output. Moreover, since stress-strain curves are time-series data, the sequence length is crucial when designing the LSTM input structure. For a sequence length $n$, the input from the $n$ sequential data points (strain and parameters) are combined to predict the $(n + 1)$-th stress. The value of the sequence length was determined through trial-and-error.

The detailed model training procedure is as follows: training data was firstly normalized before feeding into the model. The training data was then converted into tensors and structured into time-series sequences with a pre-determined sequence length. The mini-batch training with the batch sizes of 15, 8, 15, 15, 13 were used for the AlSi10Mg dataset 1, AlSi10Mg dataset 2, Ti6Al4V dataset 1, Ti6Al4V dataset 2, and carbon steel dataset, respectively. The Adam optimizer was used to optimize the hyperparameters of the models. The loss function is the mean squared error function. The models were trained over 10 epochs. In the fine-tuning stage for the training dataset in the target domain, after loading the pre-trained model, this model was re-trained with the same hyperparameter settings. Finally, the stresses were predicted based on the input from the test dataset in the target domain.

Several error metrics, including mean absolute percentage error (MAPE), root mean squared error (RMSE), and coefficient of determination ($R^2$), are used to evaluate the performance of the models. The MAPE indicates the relative error, expressed as a percentage. The RMSE is a quadratic scoring rule that measures the square root of the average of squared differences, especially sensitive for large differences. $R^2$, also known as the coefficient of determination, measures the proportion of the variance in the dependent variable that is predictable from the independent variable. These error metrics are defined as follows:

$$MAPE\ (\%) = \frac{\sum_{i=1}^{n} \frac{|y_i - \hat{y}_i|}{y_i}}{n} \times 100 \tag{9}$$

$$RMSE = \sqrt{\frac{\sum_{i=1}^{n}(y_i - \hat{y}_i)^2}{n}} \tag{10}$$

$$R^2 = 1 - \frac{\sum_{i=1}^{n}(y_i - \hat{y}_i)^2}{\sum_{i=1}^{n}(y_i - \bar{y}_i)^2} \tag{11}$$

where n is the number of data points, $y_i$ is the actual value, $\hat{y}_i$ is the predicted value, $\bar{y}_i$ is the mean of actual value.

**2.2 Dynamic time warping-based source domain selection**

Although polymers and metals differ significantly in stress-strain behavior due to their molecular structures, they share similar stress-strain behaviors, featuring a linear elastic region, followed by potential plastic deformation, yield points, and strain hardening. However, it is still challenging to quantify the similarity of the raw stress-strain curves between polymers and metals due to substantial differences in the magnitude of the stresses and strains. Therefore, the raw stress-strain curves of polymers and metals are firstly normalized. After normalization, several distance metrics can be used to measure similarity between a polymer and a metal. For example, (i) the Euclidean distance treats each curve as a long vector and sums point-by-point squared errors, which means it ignores the curve as a whole and over-penalizes even slight shifts; (ii) the Pearson correlation coefficient captures whether the curves rise and fall together but misses the shifts along the strain axis; (iii) the Fréchet distance, which measures the minimal leash length needed to traverse both curves, allows some misalignment but it is slow to compute and sensitive to outliers. By contrast, the DTW algorithm uses dynamic programming to find the optimal alignment between two sequences to compare shape similarity rather than pointwise error. In the context of this work, the DTW algorithm stretches or compresses one stress-strain curve along the strain axis to align stress-strain behaviors, which makes the DTW algorithm very effective in comparing stress-strain curves with different magnitudes.

The DTW algorithm is integrated with TL to handle non-linear alignment between the stress-strain curves of polymers and metals. Specifically, the DTW algorithm treats an entire stress-strain curve as a single functional entity to compare overall curve shape rather than independent vectors of stress and strain. The output of the DTW algorithm is a scalar distance, often referred to as the DTW distance. This distance captures how much one stress-strain curve must be warped (stretched or compressed along the strain axis) to match another. A smaller DTW distance indicates that the two stress-strain curves are more similar.

The mathematical formulation of the DTW-based source domain selection method is briefly introduced as follows [48, 50]. Suppose that we have two stress-strain curves, one is the polymer curve $P = \{(\epsilon_p^i, \sigma_p^i)\}_{i=1}^{N_p}$, where $\epsilon_p^i$ is the strain and $\sigma_p^i$ is the stress for the $i\text{-}th$ point, and $N_p$ is the number of points, the other one is the metal curve $M = \{(\epsilon_m^i, \sigma_m^i)\}_{i=1}^{N_m}$, where $\epsilon_m^i$ is the strain and $\sigma_m^i$ is the stress for the $i\text{-}th$ point, and $N_m$ is the number of points. First, to make these two curves comparable in a common scale, the strain and stress values of the polymer and metal stress-strain curves are normalized into a common range of [0,1] by dividing their corresponding maximum values, resulting in normalized curves $\tilde{P} = \{(\tilde{\epsilon}_p^i, \tilde{\sigma}_p^i)\}_{i=1}^{N_p}$ and $\tilde{M} = \{(\tilde{\epsilon}_m^i, \tilde{\sigma}_m^i)\}_{i=1}^{N_m}$.

$$\tilde{\epsilon}_p^i = \frac{\epsilon_p^i}{\max(\epsilon_p)}, \tilde{\sigma}_p^i = \frac{\sigma_p^i}{\max(\sigma_p)}, \tilde{\epsilon}_m^i = \frac{\epsilon_m^i}{\max(\epsilon_m)}, \tilde{\sigma}_m^i = \frac{\sigma_m^i}{\max(\sigma_m)} \tag{12}$$

Second, a strain grid $\{\epsilon_c^i\}_{i=1}^{N}$ is defined to down sample stress values for both curves, where $\epsilon_c^i$ is evenly spaced in [0,1] with $N$ points. Specifically, the value of $N$ is 120, therefore, the down sampled polymer curve is $P_c = \{(\tilde{\epsilon}_p^k, \tilde{\sigma}_p^k)\}_{k=1}^{120}$ and the down sampled metal curve is $M_c = \{(\tilde{\epsilon}_m^l, \tilde{\sigma}_m^l)\}_{l=1}^{120}$. Then, the local distance is obtained, where each element represents the squared difference between the $k\text{-}th$ and $l\text{-}th$ stress values from the down sampled polymer and metal curves, respectively.

$$d(k,l) = (\tilde{\sigma}_p^k - \tilde{\sigma}_m^l)^2 \tag{13}$$

Third, the cumulative cost matrix $D$ is constructed to represent the minimal cumulative cost of aligning the $k\text{-}th$ point of the polymer curve to the $l\text{-}th$ point of the metal curve. The recurrence relation is:

$$D(k,l) = d(k,l) + \min\,[\mathrm{D}(\mathrm{k}-1,l), \mathrm{D}(k,l-1), \mathrm{D}(\mathrm{k}-1,l-1)] \tag{14}$$

The recurrence relation is subject to three boundary conditions:

$$D(1,1) = d(1,1) \tag{15}$$

$$D(k,1) = d(k,1) + \mathrm{D}(\mathrm{k}-1,1)\; for\; k = 2, \ldots, 120 \tag{16}$$

$$D(1,l) = d(1,l) + \mathrm{D}(1,l-1)\; for\; l = 2, \ldots, 120 \tag{17}$$

The DTW distance is the minimum cumulative cost along a valid alignment path $\alpha$:

$$DTW(P_c, M_c) = \min_{\alpha} \sum_{(k,l)\in\alpha} D(k,l) \tag{18}$$

Here, $\alpha = [(k_1, l_1), (k_2, l_2), \ldots, (k_{120}, l_{120})]$ denotes a warping path through the matrix $D$, matching the $k\text{-}th$ point of the polymer curve to the $l\text{-}th$ point of the metal curve. A path $\alpha$ is valid if $k_{i+1} > k_i$, $l_{i+1} > l_i$, and $(k_{i+1} - k_i, l_{i+1} - l_i) \in [(1,0), (0,1), (1,1)]$. The average DTW distance between a polymer material with multiple curves $(P_1, P_2, \ldots, P_Q)$ and a metal material with multiple curves $(M_1, M_2, \ldots, M_R)$ is:

$$AvgDTW = \frac{1}{Q} \sum_{q=1}^{Q} \frac{1}{R} \sum_{r=1}^{R} DTW(P_q, M_r) \tag{19}$$

Finally, after determining the average DTW distances between each source polymer dataset and the target metal dataset, the polymer dataset with the shortest DTW distance to the target metal dataset is selected as the single optimal source domain dataset.

**2.3 Experimental design and data description**

To systematically demonstrate the effectiveness of the DTW-TL framework, several additively manufactured polymers and metals fabricated by different AM techniques were used in this study. Specifically, there were four source polymer datasets and three target metal datasets. The Nylon, PLA, and CF-ABS samples were fabricated by FFF and the Resin samples were fabricated by DLP. The AlSi10Mg and Ti6Al4V samples were fabricated by L-PBF, and the carbon steel samples were fabricated by WAAM. As summarized in Table 1, four source polymers and three target metals exhibit significantly different mechanical behaviors. Four source polymers differ not only in ultimate tensile strength (UTS) and elongation at break, but also in underlying chemistry and molecular structure. Specifically, Nylon is a semi-crystalline polyamide that can achieve an UTS of 34 MPa and an elongation at break of 210%. PLA is an amorphous polyester that can achieve an UTS of 53 MPa and an elongation at break of 6% due to its higher glass transition and limited chain mobility. CF-ABS is a high-performance thermoplastic composite combing ABS with short carbon fibers. CF-ABS can achieve an UTS of 77 MPa and an elongation at break of 1.25%. Resin is a thermoset that can achieve an UTS of 48 MPa and an elongation at break of 10%.

The Nylon, PLA, and CF-ABS samples were fabricated via an FFF-based Ultimaker S3 3D printer (Ultimaker, Netherlands) with a 250 μm nozzle diameter. 25 samples were fabricated for each polymer. The diameters of Nylon (Ultimaker, Netherlands), PLA (Pro Series, MatterHackers, USA), and CF-ABS (Push Plastic, USA) filaments were 2.85 mm. The Resin samples were fabricated via a DLP-based FlashForge Foto 3D printer (FlashForge, China) with a standard photopolymer resin. All polymer samples were in accordance with ASTM D638 Type V [51]. Uniaxial tensile tests were conducted on a universal test frame (AGS-X Series, Shimadzu, Japan) to quantitatively characterize the stress-strain curves of the samples. The load capacity of the test frame was 10 kN. The test speed and sampling frequency were 2 mm/min and 10 Hz, respectively. A full factorial design of experiments (DOE) was developed to collect the experimental data with varying parameters of each material. Table 2 and Table A.1 show the details of DOE. Each polymer involved two process parameters, and there were five numerical levels for each parameter.

Table 1. Source polymers and target metals.

| | Material | AM technique | UTS (MPa) | Elongation at break |
|---|---|---|---|---|
| Source domain (polymer) | Nylon | FFF | 34 | 210% |
| | PLA | | 53 | 6% |
| | CF-ABS | | 77 | 1.25% |
| | Resin | DLP | 48 | 10% |
| Target domain (metal) | AlSi10Mg | L-PBF | 460 | 6% |
| | Ti6Al4V | | 1060 | 15% |
| | Carbon steel | WAAM | 648 | 24% |

The AlSi10Mg samples fabricated by L-PBF solidify under extreme cooling rates, forming a fine cellular $\alpha$-Al matrix interspersed with brittle eutectic Si networks and residual micro-porosity. This microstructure delivers the lowest UTS and restricts elongation at break to 6% because the hard, interconnected Si phase and layer boundary defects act as crack initiation sites. The Ti6Al4V samples fabricated by L-PBF develop a mixed $\alpha + \beta$ structure with high lattice friction and fine prior $\beta$ grain boundaries. These features confer very high UTS while its dual phase nature still permits dislocation slip for a reasonable 15% elongation at break. The carbon steel samples fabricated by WAAM, however, cool far more slowly, yielding a coarse ferrite pearlite microstructure with larger, equiaxed grains and fewer interlayer discontinuities. Although its UTS is only intermediate of 648 MPa compared to the other metals, the coarse ferrite pearlite microstructure of carbon steel accommodates extensive plastic deformation, resulting in the highest ductility of the three metals.

The AlSi10Mg dataset was published by the Pennsylvania State University [52]. 60 AlSi10Mg samples were fabricated in accordance with ASTM E8/E8M [53] and with a 3D Systems ProX DMP 320 L-PBF system machine (3D System, Inc, United States) using gas atomized LaserForm AlSi10Mg (A) metallic powder (3D System, Inc, United States). Table 1 and Table B.1 show the DOE of the AlSi10Mg dataset. 60 AlSi10Mg samples were fabricated with varying laser powers, scanning speeds, and hatch spacings. Specifically, the samples fabricated with a hatch spacing of 0.1 mm are defined as AlSi10Mg dataset 1 (the size of the dataset is 32), the remaining samples fabricated with a hatch spacing of 0.15 mm are defined as AlSi10Mg dataset 2 (the size of the dataset is 28).

Table 2. Design of experiments for the polymers and metals.

| Domain | Material | Dataset | Parameter | Number of samples | Standard |
|---|---|---|---|---|---|
| Source domain (polymer) | Nylon | - | Print speed (mm/s)<br>Print temperature (°C) | 25 | ASTM D638 |
| | PLA | - | Print speed (mm/s)<br>Print temperature (°C) | 25 | |
| | CF-ABS | - | Print speed (mm/s)<br>Print temperature (°C) | 25 | |
| | Resin | - | UV exposure time (s)<br>Post processing time (min) | 25 | |
| Target domain (metal) | AlSi10Mg | 1 | Hatch spacing = 0.1 mm<br>Laser power (W)<br>Scanning speed (mm/s) | 32 | ASTM E8/E8M |
| | | 2 | Hatch spacing = 0.15 mm<br>Laser power (W)<br>Scanning speed (mm/s) | 28 | |
| | Ti6Al4V | 1 | Flat geometry<br>Laser power (W)<br>Scanning speed (mm/s) | 42 | |
| | | 2 | Round geometry<br>Laser power (W)<br>Scanning speed (mm/s) | 42 | |
| | Carbon steel | - | Build angle (°)<br>Nozzle angle (°) | 18 | DIN 50125 |

The Ti6Al4V dataset was also published by the Pennsylvania State University [54]. 84 Ti6Al4V samples were fabricated in accordance with the same ASTM standard and L-PBF system using LaserForm Ti Gr23(A) Ti6Al4V powder (3D Systems, Inc, United States). Table 2 and Table B.2 show the DOE of the Ti6Al4V dataset. 84 Ti6Al4V samples were fabricated with varying laser powers, scanning speeds, and geometries. Specifically, the samples fabricated with a flat geometry are defined as Ti6Al4V dataset 1 (the size of the dataset is 42), the remaining samples fabricated with a round geometry are defined as Ti6Al4V dataset 2 (the size of the dataset is 42). Uniaxial tensile tests were conducted on an electromechanical load frame (Criterion Model 45, MTS Systems, Inc., United States) with a maximum load of 10 kN and a quasi-static strain rate of $3 \times 10^{-4}$/s.

The carbon steel dataset was published by the ETH Zurich [55]. The experimental robotic welding setup consisted of an ABB IRB 4600/40 robot, a Fronius TPS 500i Pulse power source, and a Fronius 60i Robacta Drive Cold Metal Transfer (CMT) torch with a 22° neck. The standard DIN 50125 was used to design the carbon steel samples. Table 2 and Table B.3 show the DOE of the carbon steel dataset. 18 carbon steel samples were fabricated with varying build angles and nozzle angles. Uniaxial tensile tests were conducted on a Zwick universal testing machine with a capacity of 200 kN. The uniaxial tensile tests were conducted in a conditioned room at 23 °C and 50% relative humidity. The carbon steel samples were clamped on each side over a length of 40 mm. The load was applied displacement-controlled with a constant displacement rate of 0.01 mm/s.

Table 3. Train-test data split for the vanilla LSTM model without TL, the TL model trained on all four polymers as the source domain, and the DTW-TL model trained on only one polymer as the source domain.

<table>
<tr><td rowspan="7">Vanilla LSTM model without TL</td><td rowspan="2">Case</td><td rowspan="2">Training dataset</td><td rowspan="2">Test dataset</td><td colspan="2">No. of data points</td></tr>
<tr><td>Training dataset</td><td>Test dataset</td></tr>
<tr><td>1</td><td>AlSi10Mg dataset 1:<br>Sample 1 and 20</td><td>Remaining 30 samples in AlSi10Mg dataset 1</td><td>951</td><td>14,956</td></tr>
<tr><td>2</td><td>AlSi10Mg dataset 2:<br>Sample 33 and 47</td><td>Remaining 26 samples in AlSi10Mg dataset 2</td><td>443</td><td>10,240</td></tr>
<tr><td>3</td><td>Ti6Al4V dataset 1:<br>Sample 27 and 38</td><td>Remaining 40 samples in Ti6Al4V dataset 1</td><td>983</td><td>25,898</td></tr>
<tr><td>4</td><td>Ti6Al4V dataset 2:<br>Sample 69 and 80</td><td>Remaining 40 samples in Ti6Al4V dataset 2</td><td>1,778</td><td>43,180</td></tr>
<tr><td>5</td><td>Carbon steel dataset:<br>Sample 1 and 2</td><td>Remaining 16 samples in carbon steel dataset</td><td>473</td><td>3,564</td></tr>
<tr><td rowspan="7">TL model trained on all four polymers as the source domain</td><td rowspan="2">Case</td><td rowspan="2">Source domain</td><td colspan="3">Target domain</td></tr>
<tr><td>Training dataset</td><td colspan="2">Test dataset</td></tr>
<tr><td>1</td><td>Nylon, PLA,<br>CF-ABS, and Resin</td><td>AlSi10Mg dataset 1:<br>Sample 1 and 20</td><td colspan="2">Remaining 30 samples in AlSi10Mg dataset 1</td></tr>
<tr><td>2</td><td>Nylon, PLA,<br>CF-ABS, and Resin</td><td>AlSi10Mg dataset 2:<br>Sample 33 and 47</td><td colspan="2">Remaining 26 samples in AlSi10Mg dataset 2</td></tr>
<tr><td>3</td><td>Nylon, PLA,<br>CF-ABS, and Resin</td><td>Ti6Al4V dataset 1:<br>Sample 27 and 38</td><td colspan="2">Remaining 40 samples in Ti6Al4V dataset 1</td></tr>
<tr><td>4</td><td>Nylon, PLA,<br>CF-ABS, and Resin</td><td>Ti6Al4V dataset 2:<br>Sample 69 and 80</td><td colspan="2">Remaining 40 samples in Ti6Al4V dataset 2</td></tr>
<tr><td>5</td><td>Nylon, PLA,<br>CF-ABS, and Resin</td><td>Carbon steel dataset:<br>Sample 1 and 2</td><td colspan="2">Remaining 16 samples in carbon steel dataset</td></tr>
<tr><td rowspan="6">DTW-TL model trained on only one polymer as the source domain</td><td rowspan="2">Case</td><td rowspan="2">Source domain (based on DTW distance to select one)</td><td colspan="3">Target domain</td></tr>
<tr><td>Training dataset</td><td colspan="2">Test dataset</td></tr>
<tr><td>1</td><td>Nylon, or PLA, or<br>CF-ABS, or Resin</td><td>AlSi10Mg dataset 1:<br>Sample 1 and 20</td><td colspan="2">Remaining 30 samples in AlSi10Mg dataset 1</td></tr>
<tr><td>2</td><td>Nylon, or PLA, or<br>CF-ABS, or Resin</td><td>AlSi10Mg dataset 2:<br>Sample 33 and 47</td><td colspan="2">Remaining 26 samples in AlSi10Mg dataset 2</td></tr>
<tr><td>3</td><td>Nylon, or PLA, or<br>CF-ABS, or Resin</td><td>Ti6Al4V dataset 1:<br>Sample 27 and 38</td><td colspan="2">Remaining 40 samples in Ti6Al4V dataset 1</td></tr>
<tr><td>4</td><td>Nylon, or PLA, or<br>CF-ABS, or Resin</td><td>Ti6Al4V dataset 2:<br>Sample 69 and 80</td><td colspan="2">Remaining 40 samples in Ti6Al4V dataset 2</td></tr>
</table>

|  | 5 | Nylon, or PLA, or CF-ABS, or Resin | Carbon steel dataset: Sample 1 and 2 | Remaining 16 samples in carbon steel dataset |
|---|---|---|---|---|

Table 3 provides the details about the train-test data split for (1) the vanilla LSTM model without TL and (2) the TL model trained on all four polymers as the source domain as well as (3) the DTW-TL model trained on only one polymer as the source domain, respectively. To build the vanilla LSMT model without TL, only two samples in each metal dataset fabricated with the maximum and minimum values of the process parameters, respectively, were used for training. The remaining samples were used for testing. Table 3 also highlights the significant difference between the number of data points in the training and test datasets, demonstrating the effectiveness of the framework in achieving high predictive performance with limited training data. To build the TL model, all four polymer datasets in the source domain were used for pre-training, and then fine-tuned on one of the metal datasets in the target domain. For fine-tuning, the train-test data split for the TL model is the same as the train-test data split for the vanilla LSTM model without TL. All source domain datasets are first concatenated into a unified dataset and then randomly shuffled before pre-training. To build the DTW-TL model, only one single optimal polymer dataset selected by the DTW algorithm was used for pre-training, and then fine-tuned on one of the metal datasets in the target domain. For fine-tuning, the train-test data split for the DTW-TL model is the same as the train-test data split for the vanilla LSTM model without TL.

## 3. Results and discussions

### 3.1 Mechanical behaviors of additively manufactured polymers and metals

Fig. 3 shows the stress-strain curves of the Nylon, PLA, CF-ABS, Resin, AlSi10Mg, Ti6Al4V, and carbon steel samples fabricated with varying AM process parameters. Each subplot reveals the effect of varying AM process parameters on stress-strain behaviors.

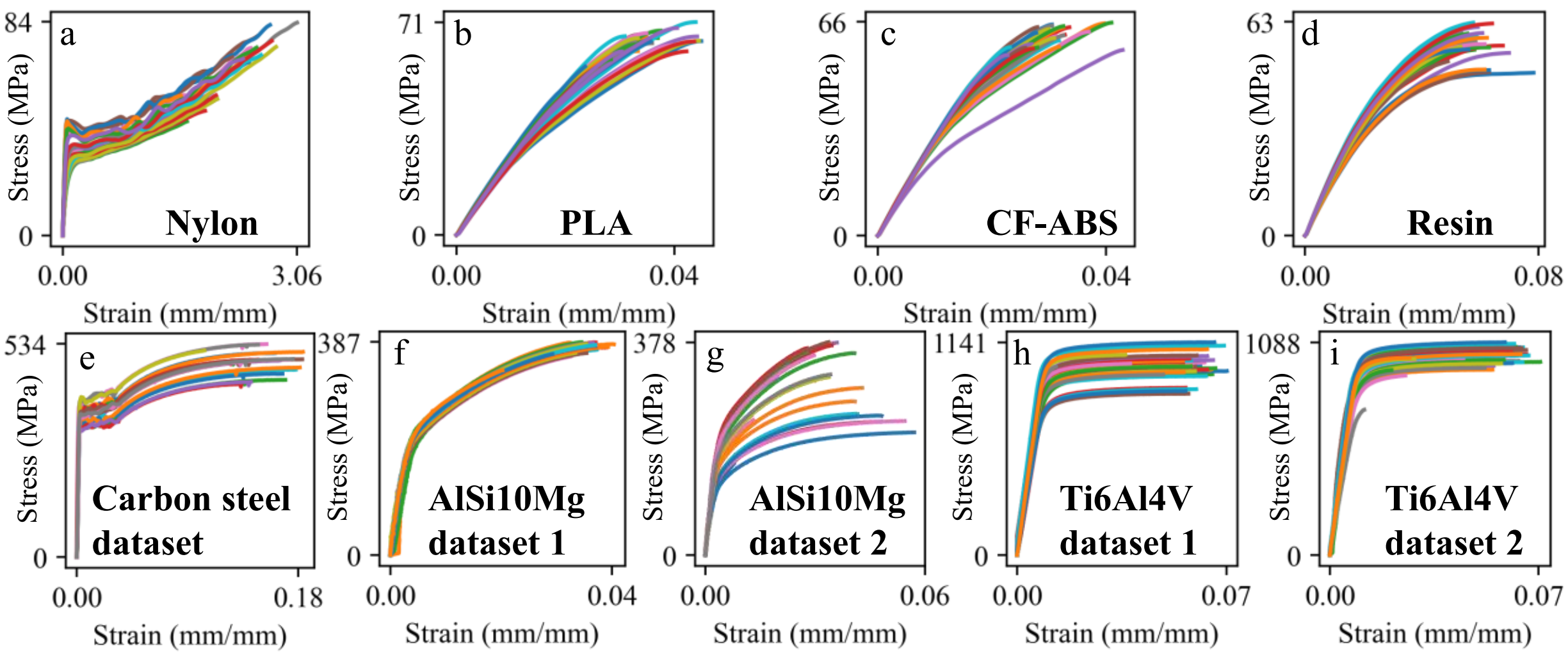

Fig. 3. Stress-strain curves of the polymer and metal samples fabricated with varying AM process parameters. Polymers: (a) Nylon, (b) PLA, and (c) CF-ABS fabricated by FFF, (d) Resin fabricated by DLP. Metals: (e) carbon steel fabricated by WAAM, (f) AlSi10Mg dataset 1, (g) AlSi10Mg dataset 2, (h) Ti6Al4V dataset 1, and (i) Ti6Al4V dataset 2 fabricated by L-PBF.

Four polymers exhibit distinct mechanical behaviors according to their stress-strain curves, including stiffness, strength, and ductility. The Nylon samples (Fig. 3a) exhibit an abrupt yield drop followed by oscillatory softening and the longest post-yield plateau region among the four polymers. The PLA (Fig. 3b) and CF-ABS samples (Fig. 3c) exhibit immediate strain hardening upon yielding. The Resin samples (Fig. 3d) exhibit a smooth, continuous post-yield plateau region. These distinct stress-strain behaviors are due to their underlying chemistry and molecular structures. Nylon is a semi-crystalline polyamide whose rigid crystalline lamellae is interspersed with an amorphous matrix. After yielding, its hydrogen-bonded chains uncoil and slide extensively, achieving the longest post-yield plateau region among the four polymers. PLA is a highly stereoregular, partially crystalline polyester, resulting in a high Young's modulus and a low elongation at break. CF-ABS combines an amorphous ABS matrix with stiff carbon fiber reinforcements. The carbon fibers increase the Young's modulus. Resin is a densely crosslinked photopolymer network. Upon yielding, covalent network bonds break and reform under load, producing a continuous post-yield plateau.

Although metals exhibit significantly higher strength and stiffness than polymers, metals and polymers share similar stress-strain behaviors, especially after yielding. Yield drops and post-yield plateau regions exhibited by the carbon steel samples (Fig. 3e) due to dislocation unlocking and band propagation in its ferrite–pearlite microstructure are similar to the Nylon samples. In contrast, both AlSi10Mg (Fig. 3f-g) and Ti6Al4V samples (Fig. 3h-i) exhibit smooth, monotonic hardening behavior, which is due to uniform slip in the $\alpha$-Al matrix with Si precipitates of AlSi10Mg samples and progressive dislocation build-up in the $\alpha + \beta$ lamellae of Ti6Al4V samples. The similarity in stress-strain behaviors between polymers and metals allows one to predict the stress-strain behaviors of additively manufactured metals by leveraging knowledge gained from a pre-trained model of the stress-strain behaviors of additively manufactured polymers.

### 3.2 Correlation between dynamic time warping distance and predictive performance

Table 4 shows the correlation between the DTW distance between a polymer dataset in the source domain and a metal dataset in the target domain and the predictive performance (i.e., MAPE) of the DTW-TL model trained on the polymer and metal datasets. The results demonstrated that the TL model trained on the polymer dataset in the source domain with the minimum DTW distance from the metal dataset in the target domain yields the best predictive performance (i.e. lowest MAPE). For AlSi10Mg datasets 1 and 2, when the Resin dataset is selected as the source domain dataset, the DTW algorithm finds the optimal alignment between their stress-strain curves (the DTW distances are 0.085 and 0.055, respectively), therefore, the DTW-TL model pre-trained on the Resin dataset and then fine-tuned on AlSi10Mg dataset 1 or 2 achieves the best predictive performance (the MAPEs are 7.01% and 17.12%, respectively). Similarly, for Ti6Al4V datasets 1 and 2, when the Resin dataset is selected as the source domain dataset, the DTW algorithm finds the optimal alignment between their stress-strain curves (the DTW distances are 0.073 and 0.061, respectively), therefore, the DTW-TL model pre-trained on the Resin dataset and then fine-tuned on Ti6Al4V dataset 1 or 2 achieves the best predictive performance (MAPEs are 11.95% and 13.09%). For the carbon steel dataset, the Nylon dataset is found to be the most similar source domain dataset with a DTW distance of 0.586. The DTW-TL model pre-trained on the Nylon dataset and then fine-tuned on carbon steel dataset achieves the lowest MAPE of 18.23%. As mentioned in Section 3.1, from the perspective of the deformation behavior of the polymers and metals, the stress-strain curves of the Resin samples are similar to the AlSi10Mg and Ti6Al4V samples; the stress-strain curves of the Nylon samples are similar to the carbon steel samples. This observation is further validated by the DTW distances. Meanwhile, high Pearson correlation coefficients between the DTW distances and the MAPEs demonstrate that the DTW distance is an effective distance metric to correlate the similarity of stress-strain behaviors and the predictive performance of the TL model.

Moreover, Fig. 4 visualizes the normalized stress-strain curves of four source polymers and three target metals after down sampling. Before applying DTW, each curve is scaled by its maximum stress and strain and resampled onto a strain grid. This pre-processing removes the effect of differences in magnitude and uneven data density, ensuring that DTW compares the shapes of stress-strain curves rather than the scale of stress-strain curves. As shown in Fig. 4a and 4b, the stress-strain curve of the Resin sample aligns the most closely with those of AlSi10Mg and Ti6Al4V samples. As shown in Fig. 4c, the stress-strain curve of the Nylon sample is the most similar to that of the carbon steel sample. These observations verify that the normalized stress-strain curves after down sampling reflect the similarity between stress-strain curves of polymers and metals.

Table 4. DTW distances and MAPEs when transferring knowledge gained from different source polymers to target metals.

| Target domain: metal | Source domain: polymer | DTW distance | MAPE | Pearson correlation coefficient between DTW distance and MAPE |
|---|---|---|---|---|
| AlSi10Mg dataset 1 | (Nylon, PLA, CF-ABS, **Resin**) | (0.173, 0.100, 0.097, **0.085**) | (42.07%, 9.80%, 9.49%, **7.01%**) | 0.996 |
| AlSi10Mg dataset 2 | (Nylon, PLA, CF-ABS, **Resin**) | (0.250, 0.063, 0.062, **0.055**) | (35.81%, 33.39%, 30.13%, **17.12%**) | 0.567 |
| Ti6Al4V dataset 1 | (Nylon, PLA, CF-ABS, **Resin**) | (0.280, 0.080, 0.079, **0.073**) | (54.76%, 16.39%, 16.24%, **11.95%**) | 0.997 |
| Ti6Al4V dataset 2 | (Nylon, PLA, CF-ABS, **Resin**) | (0.286, 0.065, 0.064, **0.061**) | (30.81%, 17.11%, 16.49%, **13.09%**) | 0.977 |
| Carbon steel dataset | (PLA, CF-ABS, Resin, **Nylon**) | (1.027, 0.999, 0.882, **0.586**) | (43.68%, 32.45%, 28.44%, **18.23%**) | 0.908 |

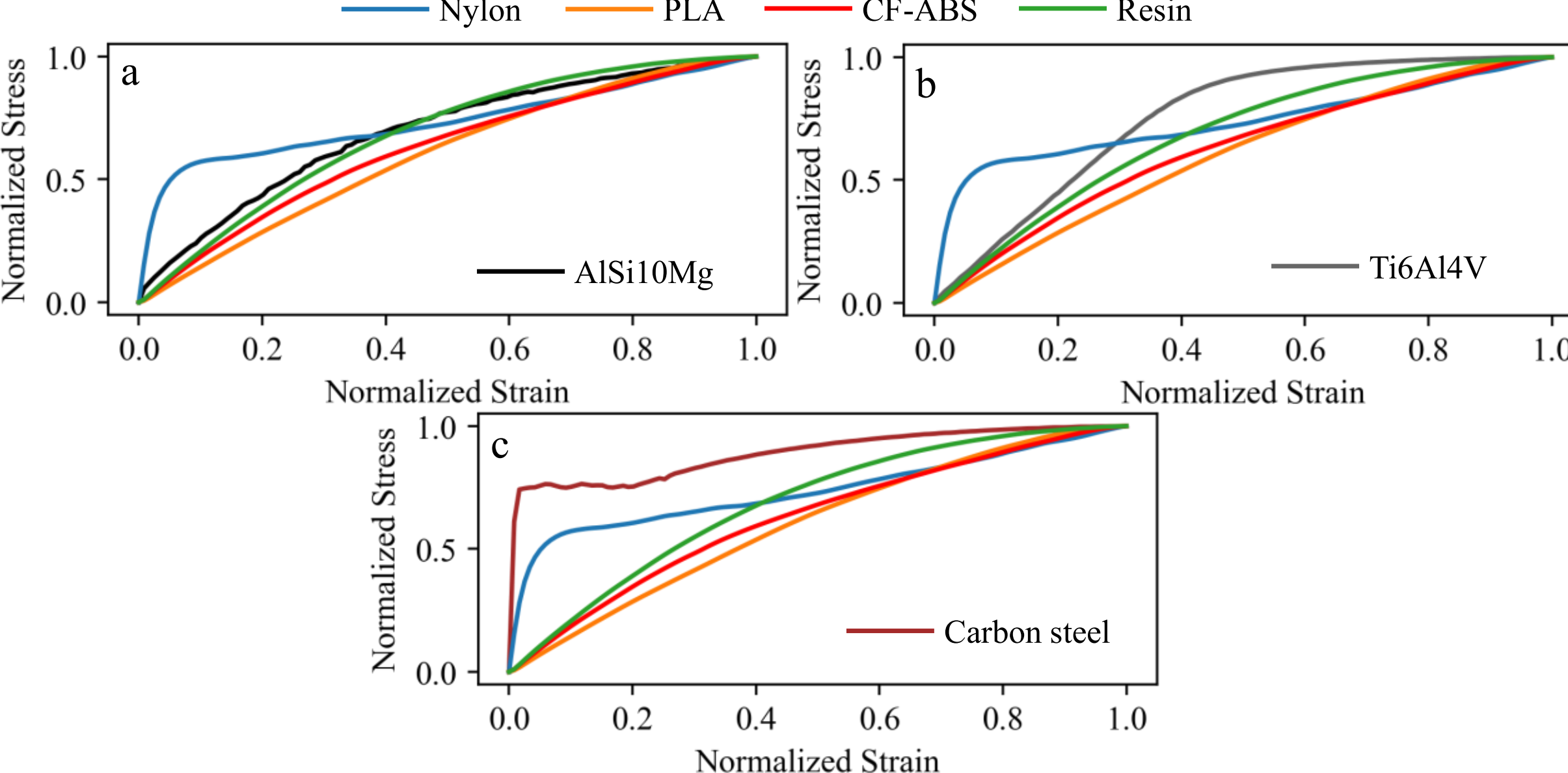

Fig. 4. Normalized and down sampled stress-strain curves of four source polymers (i.e., Nylon, PLA, CF-ABS, and Resin) compared with (a) AlSi10Mg, (b) Ti6Al4V, and (c) carbon steel.

### 3.3 Predicted stress-strain curves of AlSi10Mg

Table 5 presents the MAPEs of three models for predicting the stress-strain curves of the AlSi10Mg samples. For the vanilla LSTM model without TL, the MAPEs are relatively high, which are 13.31% for AlSi10Mg dataset 1 and 35.50% for AlSi10Mg dataset 2. The TL model trained on all four polymers as the source domain achieves better predictive performance than the vanilla LSTM model without TL. The DTW-TL model trained on only one polymer as the source domain achieves the best predictive performance among the three models by selecting one single optimal polymer dataset in the source domain with the minimum

DTW distance to AlSi10Mg datasets in the target domain. As shown in Table 5, when the Resin dataset is selected as the source domain, the DTW-TL model achieves the lowest MAPEs of 7.01% for AlSi10Mg dataset 1 and 17.12% for AlSi10Mg dataset 2. Moreover, the difference in MAPEs between AlSi10Mg dataset 1 and 2 is due to different number of data points in the training dataset and variability in stress-strain curves.

Fig. 5 shows a comparison between actual stress-strain curves of the AlSi10Mg samples and the predicted ones using different models: (1) the vanilla LSTM model without TL and (2) the TL model trained on all four polymers as the source domain as well as (3) the DTW-TL model trained on the Resin dataset as the source domain. For AlSi10Mg dataset 1, as shown in Fig. 5a, the DTW-TL model trained on the Resin dataset as the source domain outperforms other two models, especially for the region near the yield point. For AlSi10Mg dataset 2, as shown in Fig. 5b, a more noticeable difference in predictive performance among three models is observed. For example, the stress-strain curve predicted by the TL model trained on all four polymers as the source domain exhibits peaks in the elastic region which are not consistent with actual stress-strain curve. In the plastic region, the stress-strain curve predicted by the DTW-TL model effectively captures the actual stress-strain curve with high accuracy, especially in the region with higher strains, however, significant overpredictions occur for the other two models.

Table 5. MAPEs for predicting the stress-strain curves of the AlSi10Mg samples.

| Target domain | AlSi10Mg dataset 1 | AlSi10Mg dataset 2 |
|---|---|---|
| Vanilla LSTM model without TL | No source domain pre-training | |
| | 13.31% | 35.50% |
| TL model trained on all four polymers as the source domain | Source domain consists of Nylon, PLA, CF-ABS, and Resin | |
| | 8.96% | 25.04% |
| DTW-TL model trained on the Resin dataset as the source domain | Source domain only consists of **Resin** | |
| | **7.01%** | **17.12%** |

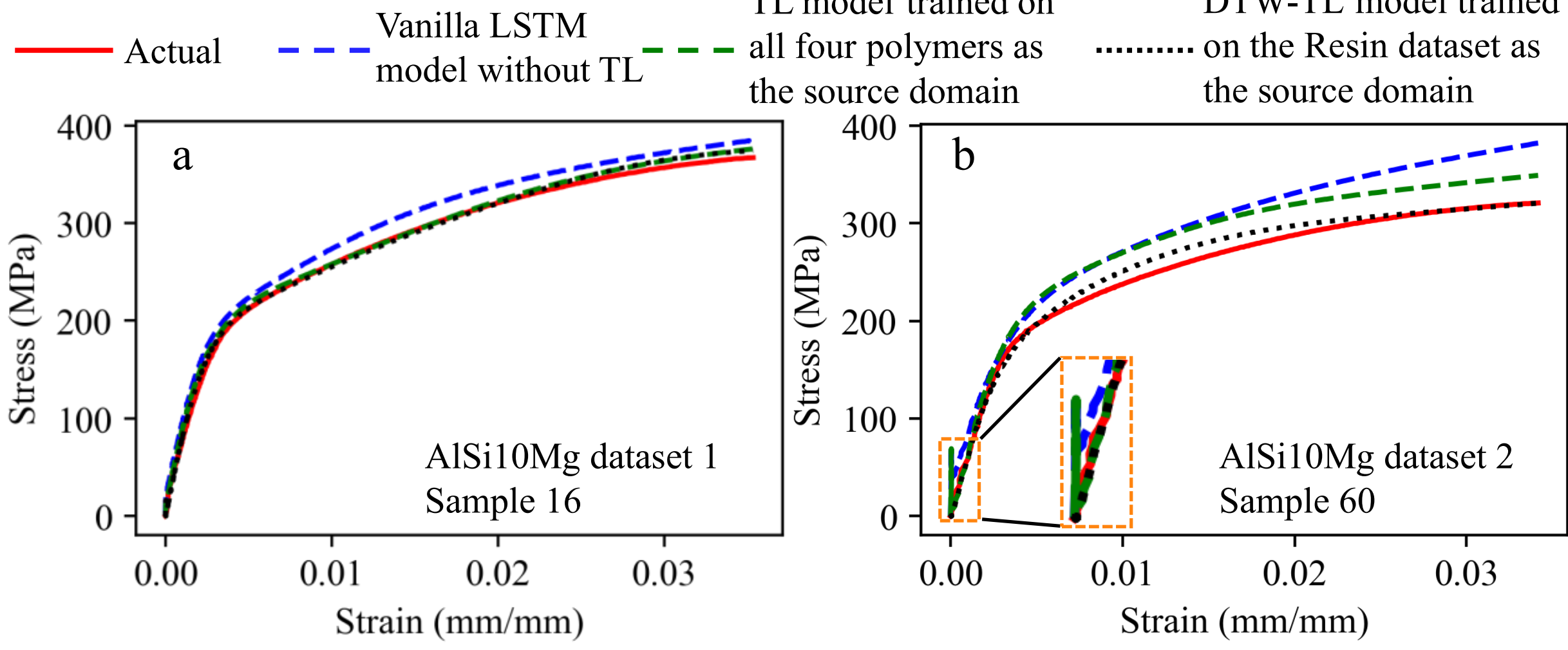

Fig. 5. Actual versus predicted stress-strain curves of AlSi10Mg samples using different models. (a) Sample 16 in AlSi10Mg dataset 1, (b) Sample 60 in AlSi10Mg dataset 2.

Fig. 6 shows the actual versus predicted stress values by the TL models for AlSi10Mg dataset 1 and 2 when different polymer datasets are used as the source domain, respectively. The red dashed line in each subplot represents the ideal case with 100% prediction accuracy, providing a benchmark for predictive performance. In addition, the violin plot in each subplot shows the distribution range of the MAPEs. For AlSi10Mg dataset 1, as shown in Fig. 6a, the TL model trained on the Nylon dataset achieves the highest average

MAPE, with a maximum value of 361%. When predicting the stress values above 200 MPa, the TL model trained on the Resin dataset achieves higher predictive performance than the TL model trained on the PLA or CF-ABS dataset. For AlSi10Mg dataset 2, similar results are observed. The TL model trained on the Nylon or CF-ABS dataset achieves the highest MAPE above 200%. The TL model trained on the Resin dataset, which achieves a relatively low MAPE range, which is between 2% and 34%. When predicting the stress values below 50 MPa, only the TL model trained on the Resin dataset achieves acceptable performance.

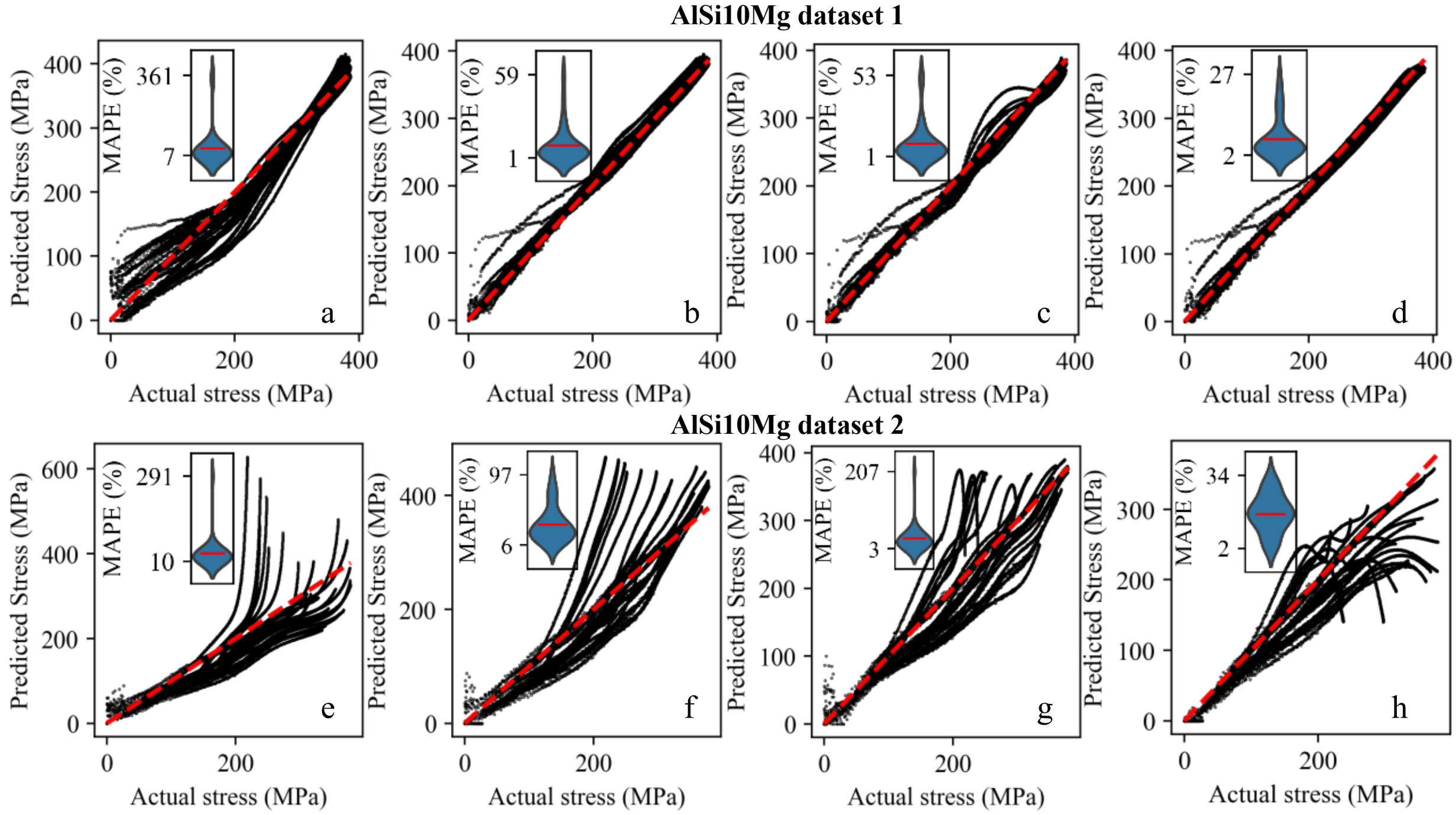

Fig. 6. Actual versus predicted stresses of AlSi10Mg samples using different polymer datasets as the source domain, respectively. (a) Nylon, (b) PLA, (c) CF-ABS, and (d) Resin for AlSi10Mg dataset 1. (e) Nylon, (f) PLA, (g) CF-ABS, and (h) Resin for AlSi10Mg dataset 2.

### 3.4 Predicted stress-strain curves of Ti6Al4V

Table 6 summarizes the MAPEs for predicting the stress-strain curves of the Ti6Al4V samples across three different models. For the vanilla LSTM model without TL, the MAPEs are relatively high, reaching 14.33% for Ti6Al4V dataset 1 and 22.82% for Ti6Al4V dataset 2. For the TL model trained on all four polymers as the source domain, the predictive performance is even worse than the vanilla LSTM model without TL. For example, the MAPEs increase from 14.33% to 18.18% for Ti6Al4V dataset 1 and from 22.82% to 30.20% for Ti6Al4V dataset 2. In contrast, by integrating the DTW algorithm to select one single optimal polymer dataset as the source domain, significant performance improvements are achieved. Specifically, when the Resin dataset is selected as the source domain, the DTW-TL model achieves the lowest MAPEs of 11.95% for Ti6Al4V dataset 1 and 13.09% for Ti6Al4V dataset 2.

Fig. 7 shows a comparison between the actual and predicted stress-strain curves of the Ti6Al4V samples using different models: (1) the vanilla LSTM model without TL and (2) the TL model trained on all four polymers as the source domain as well as (3) the DTW-TL model trained on the Resin dataset as the source domain. For Ti6Al4V dataset 1, as shown in Fig. 7a, the stress-strain curve predicted by the TL model trained on all four polymers as the source domain exhibits a noticeable peak at the beginning of the stress-strain curve. The DTW-TL model trained on the Resin dataset as the source domain outperforms the other models. For Ti6Al4V dataset 2, similar results are observed. As shown in Fig. 7b, both the vanilla LSTM model without TL and TL model trained on all four polymers as the source domain predict a peak at the beginning of the stress-strain curve. The stress-strain curve predicted by the DTW-TL model trained on the Resin dataset is in a good agreement with the actual stress-strain curve.

Table 6. MAPEs for predicting the stress-strain curves of the Ti6Al4V samples.

| Target domain | Ti6Al4V dataset 1 | Ti6Al4V dataset 2 |
| --- | --- | --- |
| Vanilla LSTM model without TL | No source domain pre-training | |
| | 14.33% | 22.82% |
| TL model trained on all four polymers as the source domain | Source domain consists of Nylon, PLA, CF-ABS, and Resin | |
| | 18.18% | 30.20% |
| DTW-TL model trained on the Resin dataset as the source domain | Source domain consists of only **Resin** | |
| | **11.95%** | **13.09%** |

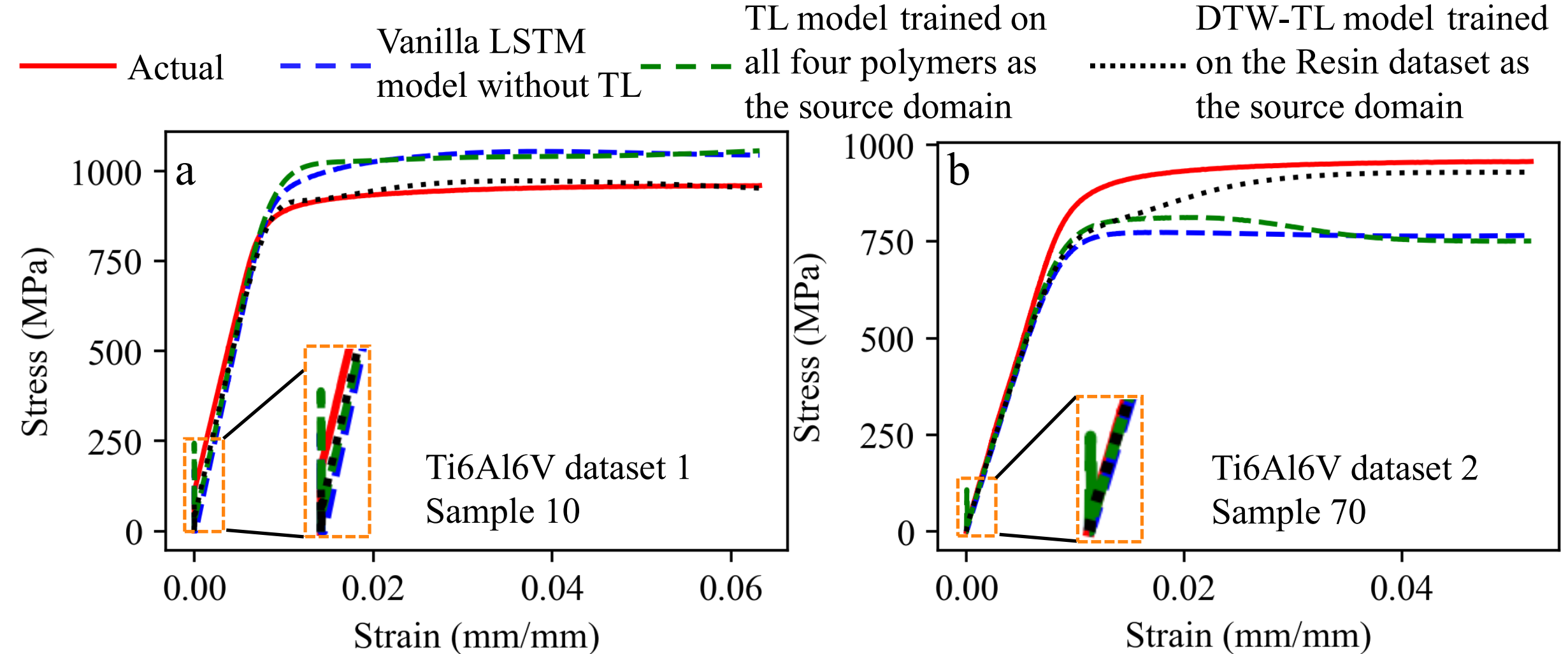

Fig. 7. Actual versus predicted stress-strain curves of Ti6Al4V samples using different models. (a) Sample 10 in Ti6Al4V dataset 1, (b) Sample 70 in Ti6Al4V dataset 2.

Fig. 8 shows the actual and predicted stress values by the TL models for Ti6Al4V dataset 1 and 2 when different polymer datasets are used as the source domain, respectively. For Ti6Al4V dataset 1, predictions of the TL model trained on the Nylon dataset (Fig. 8a) achieve the highest average MAPE. The scatter plot reveals that the predictions of the TL model trained on the Nylon dataset show substantial deviations from the actual stress values in the entire stress-strain curve, while the TL trained on the PLA dataset (Fig. 8b) performs poorly in the stress range between 750 MPa and 1000 MPa. For the predicted stress values below 50 MPa, when using the Resin dataset as the source domain (Fig. 8d), the predicted and actual stress values are in a very good agreement. For Ti6Al4V dataset 2, similar results are observed. The TL model trained on the Nylon (Fig. 8e) or PLA (Fig. 8f) dataset achieve a larger range of MAPEs. Compared with the TL model trained on the CF-ABS (Fig. 8g) dataset, the TL model trained on the Resin (Fig. 8h) dataset achieves the lowest MAPEs ranging from 4% to 26%.

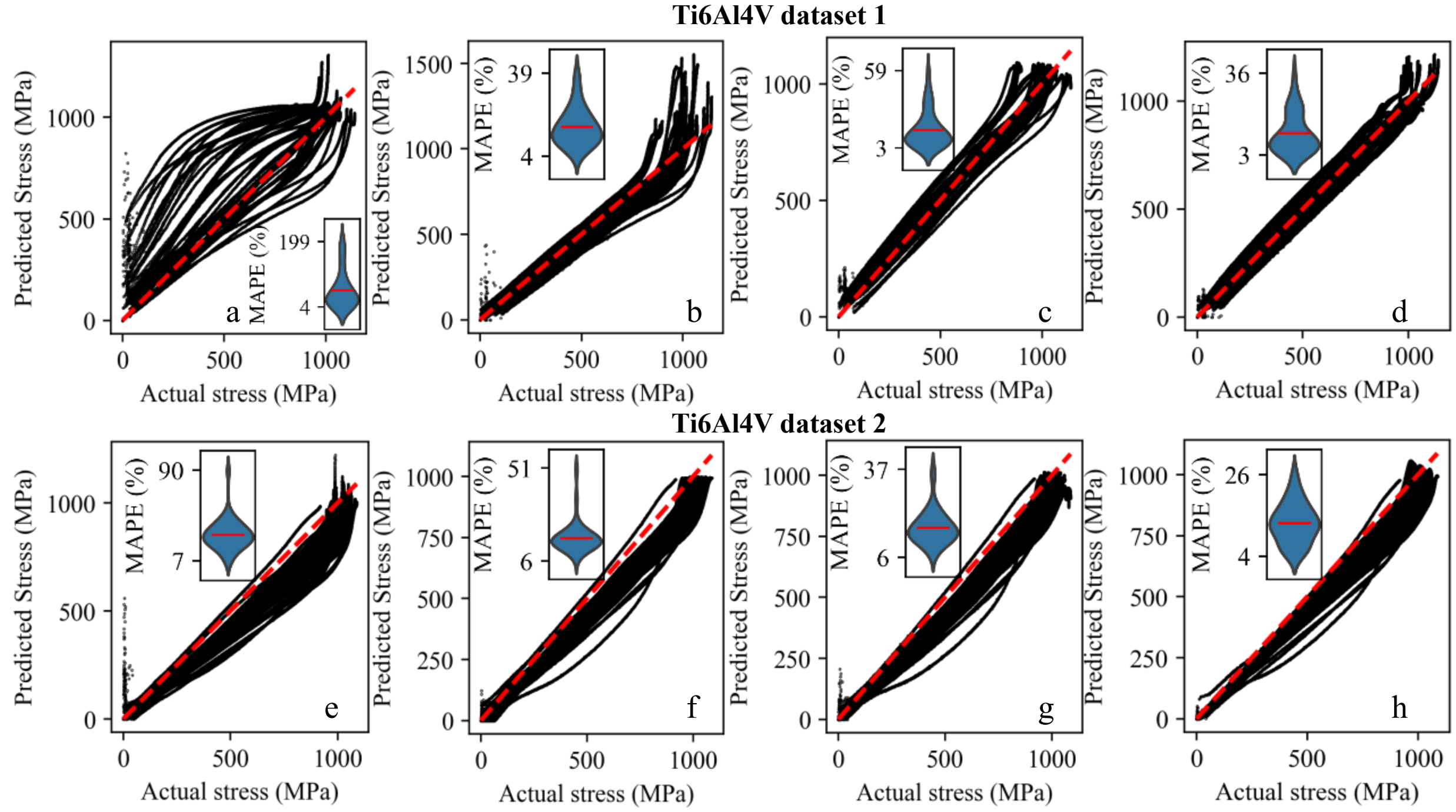

Fig. 8. Actual versus predicted stresses of Ti6Al4V samples using different polymer datasets as the source domain, respectively. (a) Nylon, (b) PLA, (c) CF-ABS, and (d) Resin for Ti6Al4V dataset 1. (e) Nylon, (f) PLA, (g) CF-ABS, and (h) Resin for Ti6Al4V dataset 2.

### 3.5 Predicted stress-strain curves of carbon steel

Table 7 provides the MAPEs for predicting the stress-strain curves of the carbon steel samples across three different models. The vanilla LSTM model without TL achieves a relatively high MAPE of 26.28%. The TL model trained on all four polymers as the source domain achieves a MAPE of 21.25%. The DTW-TL model trained on the Nylon dataset as the source domain achieves a MAPE of 18.23%.

Table 7. MAPEs for predicting the stress-strain curves of the carbon steel samples.

| Target domain | Carbon steel dataset |
| --- | --- |
| Vanilla LSTM model without TL | No source domain pre-training |
| | 26.28% |
| TL model trained on all four polymers as the source domain | Source domain consists of Nylon, PLA, CF-ABS, and Resin |
| | 21.25% |
| DTW-TL model trained on the Nylon dataset as the source domain | Source domain consists of only **Nylon** |
| | **18.23%** |

Fig. 9 shows a comparison between the actual and predicted stress-strain curves of the carbon steel samples using three different models: (1) the vanilla LSTM model without TL and (2) the TL model trained on all four polymers as the source domain as well as (3) the DTW-TL model trained on the Nylon dataset as the source domain. Within the region before yielding, all three models achieve high prediction accuracy. However, the DTW-TL trained on the Nylon dataset as the source domain outperforms the other two models within the post-yield region.

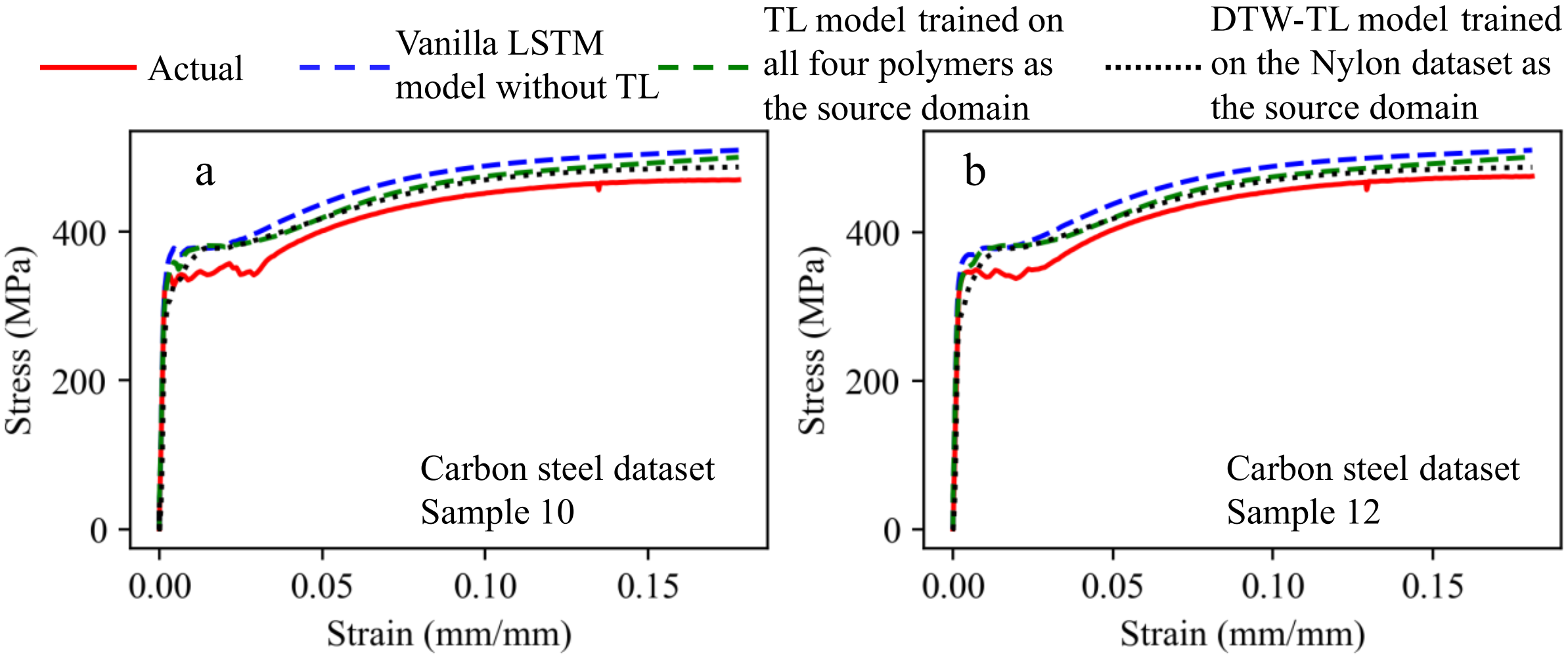

Fig. 9. Actual versus predicted stress-strain curves of carbon steel samples using different models. (a) Sample 10 and (b) Sample 12 in carbon steel dataset.

Fig. 10 shows the actual and predicted stress values by the TL models for the carbon steel samples when different polymer datasets are used as the source domain, respectively. When using the PLA (Fig. 10a) or CF-ABS (Fig. 10b) dataset as the source domain, the predicted stress values by the TL model exhibits noticeable deviations from the actual ones for the entire stress-strain curve, yielding a relatively high MAPE range from 15% to 106% and 15% to 73%, respectively. The MAPEs of the TL model trained on the Resin (Fig. 10c) dataset range between 13% and 54%. In contrast, the TL model trained on the Nylon dataset ( Fig. 10d) performs the best with MAPEs ranging between 8% and 34%.

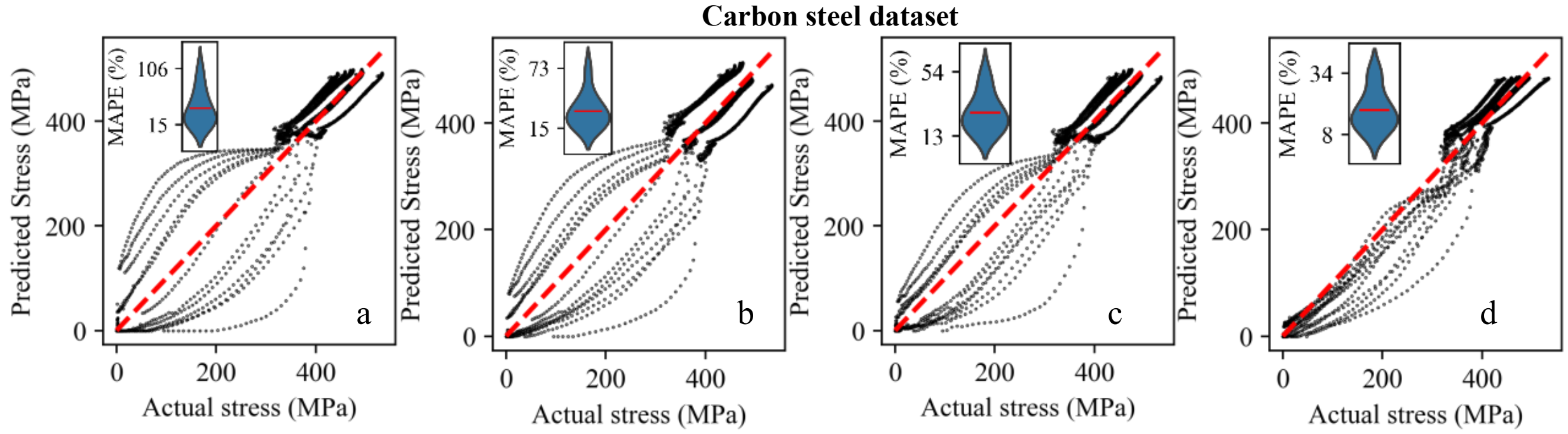

Fig. 10. Actual versus predicted stresses of carbon steel samples using different polymer datasets as the source domain, respectively. (a) PLA, (b) CF-ABS, (c) Resin, and (d) Nylon.

### 3.6 Predictive performance comparison

Table 8 provides a comparison of the average predictive performance of the vanilla LSTM model without TL, the TL model trained on all four polymer datasets, and the DTW-TL model trained on only one polymer dataset across all target domain metal datasets, including AlSi10Mg dataset 1, AlSi10Mg dataset 2, Ti6Al4V dataset 1, Ti6Al4V dataset 2, and carbon steel dataset. Interestingly, the vanilla LSTM model without TL outperforms the TL model trained on all four polymers as the source domain. Specifically, the TL model trained on all four polymers as the source domain achieves a MAPE of 22.92%, whereas the vanilla LSTM model without TL achieves a MAPE of 20.52%. This suggests that without optimal domain alignment, simply increasing the size and diversity of the source domain dataset can negatively impact predictive performance. In contrast, the DTW-TL model trained on the Resin or Nylon dataset as the source domain achieves the best predictive performance with a MAPE of 12.41%, a RMSE of 63.75, and a $R^2$ of 0.96, although there are only 25 samples in the source domain dataset. These results demonstrated the effectiveness of the DTW-TL framework by selecting the optimal polymer dataset in the source domain that aligns with the metal dataset in the target domain.

Table 8. Comparison of the average predictive performance.

| | MAPE | RMSE | R2 | No. of samples in source domain dataset |
|---|---|---|---|---|
| Vanilla LSTM model without TL | No source domain pre-training | | | |
| | 20.52% | 81.55 | 0.94 | 0 |
| TL model trained on all four polymers as the source domain | Source domain consists of Nylon, PLA, CF-ABS, and Resin | | | |
| | 22.92% | 74.45 | 0.94 | 100 |
| DTW-TL model trained on one polymer dataset as the source domain | Source domain consists of only **Nylon or Resin** | | | |
| | **12.41%** | **63.75** | **0.96** | **25** |

## 4. Conclusions

A novel DTW-TL framework for AM part qualification was developed to predict the complex stress-strain behaviors of additively manufactured high-performance, expensive metals by leveraging the knowledge gained from additively manufactured low-cost polymers. The DTW algorithm was used to select one single optimal polymer dataset among multiple polymer datasets in the source domain. The optimal polymer exhibits the most similar stress-strain behavior to the metal in the target domain. The framework was demonstrated on three metal datasets in the target domain. The framework was demonstrated on three metals in the target domain under tensile loading, including the AlSi10Mg, Ti6Al4V, and carbon steel datasets. When the target domain contains the AlSi10Mg or Ti6Al4V dataset, the Resin dataset was selected as the optimal source domain because the DTW algorithm revealed that the AlSi10Mg, Ti6Al4V, and Resin samples exhibit a continuous post-yield plateau region, respectively. When the carbon steel dataset was selected as the target domain, the Nylon dataset was selected as the optimal source domain because the DTW algorithm revealed that both Nylon and carbon steel samples exhibit yield drops and subsequent plateau regions. The experimental results showed that the DTW-TL model trained on one single optimal polymer dataset in the source domain achieved the lowest average MAPE of 12.41%, the lowest RMSE of 63.75, and the highest $R^2$ of 0.96 across three target metal datasets, outperforming the vanilla LSTM model without TL and the TL model trained on all four polymer datasets in the source domain. Notably, this performance improvement was achieved using a substantially smaller source domain dataset of 25 samples compared to the initial source domain dataset with 100 samples. In addition, pre-training the TL model on all four polymer datasets did not improve predictive performance, instead, it resulted in a greater MAPE of 22.92% compared to the vanilla LSTM model without TL. These results demonstrated the effectiveness of the DTW-TL framework in selecting one single optimal source domain dataset as well as transferring knowledge in stress-strain behaviors from additively manufactured low-cost polymers to high-performance, expensive additively manufactured metals. In the future, we will demonstrate the DTW-TL framework on a wider range of additively manufactured polymers and metals.

**Acknowledgments**

This research is sponsored by the Defense Advanced Research Projects Agency. The content of the information does not necessarily reflect the position or the policy of the Government. No official endorsement should be inferred. We would also like to thank Dr. Qingyang Liu for collecting the raw data of four polymer datasets.

**Data Availability Statement**

The dataset generated and supporting the findings of this article are available upon reasonable request from the corresponding author.

**Appendix A. Process parameters for additively manufactured polymer samples**

Table A.1 Process parameters for Nylon, PLA, and CF-ABS samples.

| Nylon | | | PLA | | | CF-ABS | | |
|---|---|---|---|---|---|---|---|---|
| Sample ID | Print temperature (°C) | Print speed (mm/s) | Sample ID | Print temperature (°C) | Print speed (mm/s) | Sample ID | Print temperature (°C) | Print speed (mm/s) |
| 1 | 220 | 10 | 1 | 180 | 10 | 1 | 200 | 10 |
| 2 | 220 | 20 | 2 | 180 | 30 | 2 | 200 | 30 |
| 3 | 220 | 30 | 3 | 180 | 50 | 3 | 200 | 50 |
| 4 | 220 | 40 | 4 | 180 | 70 | 4 | 200 | 70 |
| 5 | 220 | 50 | 5 | 180 | 90 | 5 | 200 | 90 |
| 6 | 230 | 10 | 6 | 200 | 10 | 6 | 220 | 10 |
| 7 | 230 | 20 | 7 | 200 | 30 | 7 | 220 | 30 |
| 8 | 230 | 30 | 8 | 200 | 50 | 8 | 220 | 50 |
| 9 | 230 | 40 | 9 | 200 | 70 | 9 | 220 | 70 |
| 10 | 230 | 50 | 10 | 200 | 90 | 10 | 220 | 90 |
| 11 | 240 | 10 | 11 | 220 | 10 | 11 | 240 | 10 |
| 12 | 240 | 20 | 12 | 220 | 30 | 12 | 240 | 30 |
| 13 | 240 | 30 | 13 | 220 | 50 | 13 | 240 | 50 |
| 14 | 240 | 40 | 14 | 220 | 70 | 14 | 240 | 70 |
| 15 | 240 | 50 | 15 | 220 | 90 | 15 | 240 | 90 |
| 16 | 250 | 10 | 16 | 240 | 10 | 16 | 260 | 10 |
| 17 | 250 | 20 | 17 | 240 | 30 | 17 | 260 | 30 |
| 18 | 250 | 30 | 18 | 240 | 50 | 18 | 260 | 50 |
| 19 | 250 | 40 | 19 | 240 | 70 | 19 | 260 | 70 |
| 20 | 250 | 50 | 20 | 240 | 90 | 20 | 260 | 90 |
| 21 | 260 | 10 | 21 | 260 | 10 | 21 | 280 | 10 |
| 22 | 260 | 20 | 22 | 260 | 30 | 22 | 280 | 30 |
| 23 | 260 | 30 | 23 | 260 | 50 | 23 | 280 | 50 |
| 24 | 260 | 40 | 24 | 260 | 70 | 24 | 280 | 70 |
| 25 | 260 | 50 | 25 | 260 | 90 | 25 | 280 | 90 |

Table A.1 Process parameters for the Resin samples.

| Sample ID | UV exposure time (s) | Post processing time (min) |
|---|---|---|
| 1 | 2 | 2 |
| 2 | 2 | 4 |
| 3 | 2 | 6 |
| 4 | 2 | 8 |
| 5 | 2 | 10 |
| 6 | 2.5 | 2 |
| 7 | 2.5 | 4 |
| 8 | 2.5 | 6 |
| 9 | 2.5 | 8 |
| 10 | 2.5 | 10 |
| 11 | 3 | 2 |
| 12 | 3 | 4 |
| 13 | 3 | 6 |
| 14 | 3 | 8 |
| 15 | 3 | 10 |
| 16 | 3.5 | 2 |
| 17 | 3.5 | 4 |
| 18 | 3.5 | 6 |
| 19 | 3.5 | 8 |
| 20 | 3.5 | 10 |
| 21 | 4 | 2 |
| 22 | 4 | 4 |
| 23 | 4 | 6 |
| 24 | 4 | 8 |
| 25 | 4 | 10 |

**Appendix B. Process parameters for the additively manufactured metal samples**

Table B.1 Process parameters for the AlSi10Mg samples.

| Dataset 1: hatch spacing = 0.1mm | | | Dataset 2: hatch spacing = 0.15mm | | |
|---|---|---|---|---|---|
| Sample ID | Laser power (W) | Scanning speed (mm/s) | Sample ID | Laser power (W) | Scanning speed (mm/s) |
| 1 | 60 | 250 | 33 | 60 | 250 |
| 2 | 160 | 250 | 34 | 160 | 250 |
| 3 | 160 | 800 | 35 | 160 | 800 |
| 4 | 160 | 1350 | 36 | 260 | 250 |
| 5 | 260 | 250 | 37 | 260 | 800 |
| 6 | 260 | 800 | 38 | 260 | 1350 |
| 7 | 260 | 1350 | 39 | 360 | 250 |
| 8 | 260 | 1900 | 40 | 360 | 800 |
| 9 | 360 | 250 | 41 | 360 | 1350 |
| 10 | 360 | 800 | 42 | 360 | 1900 |
| 11 | 360 | 1350 | 43 | 460 | 250 |
| 12 | 360 | 1900 | 44 | 460 | 800 |
| 13 | 360 | 2450 | 45 | 460 | 1350 |
| 14 | 360 | 3000 | 46 | 460 | 1900 |
| 15 | 460 | 250 | 47 | 460 | 2450 |
| 16 | 460 | 800 | 48 | 110 | 250 |
| 17 | 460 | 1350 | 49 | 110 | 500 |
| 18 | 460 | 1900 | 50 | 160 | 500 |
| 19 | 460 | 2450 | 51 | 260 | 500 |
| 20 | 460 | 3000 | 52 | 360 | 500 |
| 21 | 110 | 250 | 53 | 460 | 500 |
| 22 | 210 | 250 | 54 | 210 | 250 |
| 23 | 60 | 500 | 55 | 210 | 500 |
| 24 | 110 | 500 | 56 | 210 | 800 |
| 25 | 160 | 500 | 57 | 210 | 1100 |
| 26 | 210 | 500 | 58 | 260 | 1100 |
| 27 | 260 | 500 | 59 | 360 | 1100 |
| 28 | 360 | 500 | 60 | 460 | 1100 |
| 29 | 460 | 500 | | | |

| | | | |
|---|---|---|---|
| 30 | 110 | 800 | |
| 31 | 210 | 800 | |
| 32 | 210 | 1350 | |

Table B.2 Process parameters for the Ti6Al4V samples.

| Dataset 1: flat geometry | | | Dataset 2: round geometry | | |
|---|---|---|---|---|---|
| Sample ID | Laser power (W) | Scanning speed (mm/s) | Sample ID | Laser power (W) | Scanning speed (mm/s) |
| 1 | 275 | 800 | 43 | 275 | 800 |
| 2 | 275 | 760 | 44 | 275 | 760 |
| 3 | 275 | 720 | 45 | 275 | 720 |
| 4 | 275 | 680 | 46 | 275 | 680 |
| 5 | 275 | 640 | 47 | 275 | 640 |
| 6 | 275 | 600 | 48 | 275 | 600 |
| 7 | 275 | 840 | 49 | 275 | 840 |
| 8 | 275 | 880 | 50 | 275 | 880 |
| 9 | 275 | 920 | 51 | 275 | 920 |
| 10 | 275 | 960 | 52 | 275 | 960 |
| 11 | 275 | 1000 | 53 | 275 | 1000 |
| 12 | 175 | 800 | 54 | 175 | 800 |
| 13 | 195 | 800 | 55 | 195 | 800 |
| 14 | 215 | 800 | 56 | 215 | 800 |
| 15 | 235 | 800 | 57 | 235 | 800 |
| 16 | 255 | 800 | 58 | 255 | 800 |
| 17 | 295 | 800 | 59 | 295 | 800 |
| 18 | 315 | 800 | 60 | 315 | 800 |
| 19 | 335 | 800 | 61 | 335 | 800 |
| 20 | 355 | 800 | 62 | 355 | 800 |
| 21 | 375 | 800 | 63 | 375 | 800 |
| 22 | 135 | 400 | 64 | 135 | 400 |
| 23 | 205 | 600 | 65 | 205 | 600 |
| 24 | 345 | 1000 | 66 | 345 | 1000 |
| 25 | 415 | 1200 | 67 | 415 | 1200 |
| 26 | 485 | 1400 | 68 | 485 | 1400 |
| 27 | 500 | 1480 | 69 | 500 | 1480 |
| 28 | 155 | 400 | 70 | 155 | 400 |
| 29 | 235 | 600 | 71 | 235 | 600 |
| 30 | 395 | 1000 | 72 | 395 | 1000 |
| 31 | 475 | 1200 | 73 | 475 | 1200 |
| 32 | 500 | 1290 | 74 | 500 | 1290 |
| 33 | 115 | 400 | 75 | 115 | 400 |
| 34 | 175 | 600 | 76 | 175 | 600 |
| 35 | 295 | 1000 | 77 | 295 | 1000 |
| 36 | 355 | 1200 | 78 | 355 | 1200 |
| 37 | 475 | 1600 | 79 | 475 | 1600 |
| 38 | 75 | 400 | 80 | 75 | 400 |
| 39 | 115 | 600 | 81 | 115 | 600 |
| 40 | 195 | 1000 | 82 | 195 | 1000 |
| 41 | 235 | 1200 | 83 | 235 | 1200 |
| 42 | 315 | 1600 | 84 | 315 | 1600 |

Table B.3 Process parameters for the carbon steel samples.

| Sample ID | Build angle (°) | Nozzle angle (°) |
| --- | --- | --- |
| 1 | 0 | 0 |
| 2 | 0 | 0 |
| 3 | 0 | 0 |
| 4 | 0 | 22.5 |
| 5 | 0 | 22.5 |
| 6 | 0 | 22.5 |
| 7 | 0 | 45 |
| 8 | 0 | 45 |
| 9 | 0 | 45 |
| 10 | 45 | 0 |
| 11 | 45 | 0 |
| 12 | 45 | 0 |
| 13 | 45 | 22.5 |
| 14 | 45 | 22.5 |
| 15 | 45 | 22.5 |
| 16 | 45 | 45 |
| 17 | 45 | 45 |
| 18 | 45 | 45 |